\documentclass[letterpaper, 10 pt, conference]{ieeeconf}  % Comment this line out if you need a4paper

\IEEEoverridecommandlockouts                              % This command is only needed if 
\let\labelindent\relax
\usepackage{graphics}
\usepackage{amsfonts}
\usepackage{amsmath}
\usepackage{bm}
\usepackage{upgreek}

\usepackage{amssymb}
\usepackage{leftindex}
\usepackage{algorithm}
\usepackage{algpseudocode}
\usepackage{epsfig}
\usepackage{graphicx}
\usepackage{svg}
\usepackage{caption}
\usepackage{mathrsfs}
\usepackage{array}
\usepackage{multirow}
\usepackage{multicol}
\usepackage{booktabs}
\usepackage{makecell}
\usepackage{xcolor}
\usepackage{amsfonts}
\usepackage{flushend}
\usepackage{soul}
\usepackage{hyperref}
\hypersetup{colorlinks,linkcolor=black}
\usepackage{stfloats}
\usepackage{balance}
\usepackage{threeparttable}
\usepackage{enumitem}
\usepackage{bbding}

\newcommand\norm[1]{\lVert #1 \rVert}

\title{\LARGE \bf
CNSv2: Probabilistic Correspondence Encoded Neural Image Servo
}

\author{Anzhe Chen, Hongxiang Yu, Shuxin Li, Yuxi Chen, Zhongxiang Zhou, Wentao Sun, Rong Xiong, Yue Wang$^{*}$% <-this % stops a space
}

\begin{document}

\maketitle
\thispagestyle{empty}
\pagestyle{empty}

%%%%%%%%%%%%%%%%%%%%%%%%%%%%%%%%%%%%%%%%%%%%%%%%%%%%%%%%%%%%%%%%%%%%%%%%%%%%%%%%
\begin{abstract}

Visual servo based on traditional image matching methods often requires accurate keypoint correspondence for high precision control. However, keypoint detection or matching tends to fail in challenging scenarios with inconsistent illuminations or textureless objects, resulting significant performance degradation. Previous approaches, including our proposed Correspondence encoded Neural image Servo policy (CNS), attempted to alleviate these issues by integrating neural control strategies. While CNS shows certain improvement against error correspondence over conventional image-based controllers, it could not fully resolve the limitations arising from poor keypoint detection and matching. In this paper, we continue to address this problem and propose a new solution: Probabilistic Correspondence Encoded Neural Image Servo (CNSv2). CNSv2 leverages probabilistic feature matching to improve robustness in challenging scenarios. By redesigning the architecture to condition on multimodal feature matching, CNSv2 achieves high precision, improved robustness across diverse scenes and runs in real-time. We validate CNSv2 with simulations and real-world experiments, demonstrating its effectiveness in overcoming the limitations of detector-based methods in visual servo tasks.

\end{abstract}

%%%%%%%%%%%%%%%%%%%%%%%%%%%%%%%%%%%%%%%%%%%%%%%%%%%%%%%%%%%%%%%%%%%%%%%%%%%%%%%%

\section{Introduction}
\label{sec:intro}

Visual servo is an essential technique in robotics which enables precise relocalization. Traditional methods \cite{chaumette2006visual} including image-based visual servo (IBVS), position-based visual servo (PBVS) and hybrid approaches require accurate keypoint correspondence between current and desired images to estimate image Jacobian or camera pose for control. These methods adopt explicit correspondence as input abstraction, which generalize well in novel scenes, but is sensitive to matching errors. To overcome the matching problem in challenging scenes, several learning based approaches are proposed \cite{saxena2017exploring,bateux2018training,yu2019siamese,felton2021siame}, which bypass the matching and directly use the implicit image features for control. While these methods converge well and achieve high precision in trained scenes, however, cannot generalize well to novel scenes. In our previous work, CNS \cite{chen2024cns}, we still adopt the explicit correspondence as inputs but tackle non-idealities with a graph neural network based policy.

CNS achieves high success ratio and precision in textured scenes with strong generalization capabilities, however, facing challenges in textureless scenes or large illumination variations. These limitations arise from its reliance on a detector-based image matching frontend, which struggles to detect keypoints in textureless scenes and is not robust to illumination variations.

Recently, detector-free image matching approaches \cite{sun2021loftr,edstedt2023dkm,edstedt2024RoMa} have shown robust matching in challenging conditions. Leveraging multimodalities in the coarsest feature correspondence, they can predict matches in textureless region and achieves certain robustness to illumination variations. However, CNS is intrinsically incompatible with these detector-free methods as it requires static desired keypoints. If we use these matches for IBVS/PBVS control, we reduce the probabilistic matching to explicit matching, losing the opportunity to correct the inaccurate matches with multimodalities. Moreover, the performance is again limited by their sensitivity to error matches. This brings up a question: \textit{Can we utilize the multimodal probabilistic matching combined with CNS for visual servoing?}

\begin{figure}[t]
\includegraphics[width=\linewidth]{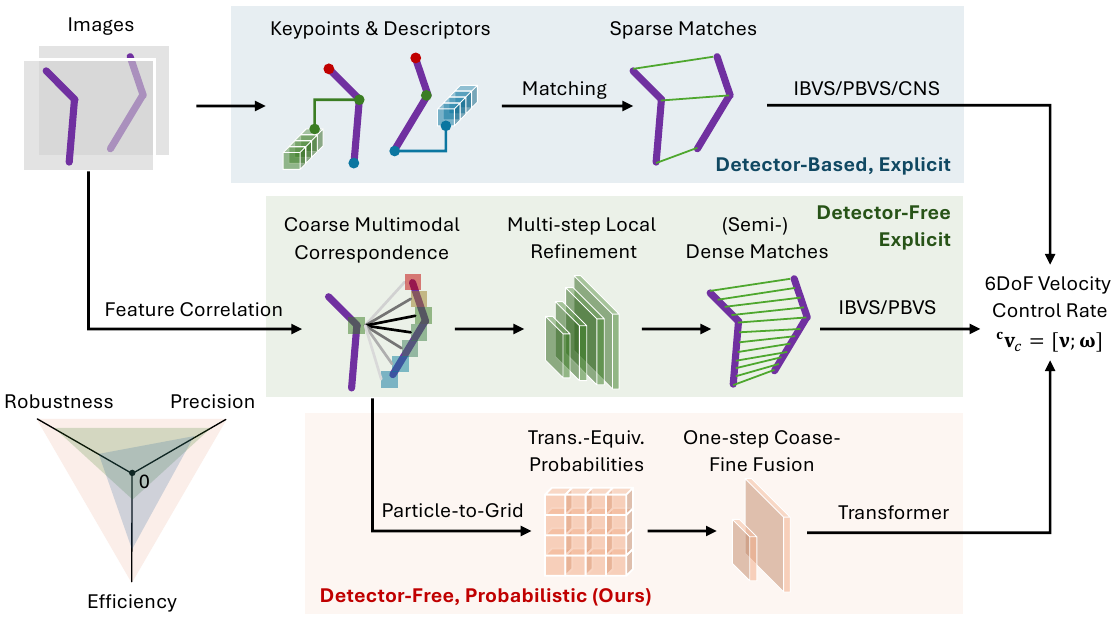}
\centering
\caption{We utilize probabilistic correspondence of robust features from foundation model and use neural policy for control, endowing image servo with generalization, high precision and robustness to challenging scenes.}
\label{fig: teaser}
  \vspace{-0.5cm}
\end{figure}

A straightforward idea is to use the fine-grained features from the last layer of methods like RoMa \cite{edstedt2024RoMa}, along with a multilayer perceptron (MLP), to predict velocity control. However, achieving a high convergence, precision, and real-time visual servo policy poses two significant challenges: (1) Although detector-free methods offer multimodal matching, their multistep refinement results in low inference efficiency. How can we balance inference efficiency, precision, and convergence? (2) Visual servo task predicts low dimensional velocity, the supervision is weaker than image matching task that has pixel-wise dense supervision, which makes training difficult. Considering that a general visual servo policy also needs to cope with variations of camera intrinsics and scene scales, the training becomes harder.

In this paper, we propose CNSv2, which maintains the generalization ability of CNS while leveraging the multimodal matching for visual servo.
To address the aforementioned challenges, we first propose a translation-equivariant representation of probabilistic matching that eases the learning complexity. Additionally, we decouple the network predictions from real-world scales and camera intrinsics, simplifying the training and reducing the data requirements. Second, we incorporate foundation models to provide consistent and robust coarse features. In this way, we only retain one global correlation layer, bypassing the computationally expensive multistep local refinement of current detector-free methods, and instead use only several low-cost convolution layers to fuse fine-grained features, which significantly improves the model efficiency. Furthermore, we derive a hybrid control strategy that optimizes both Cartesian and image trajectories for better convergence. In addition, we employ mixed floating point training and inference to ensures the model runs in real-time. Finally, CNSv2 arrives at a general visual servo policy with high convergence, efficiency and precision. Our contributions are threefolds:
\begin{itemize} %[leftmargin=*]
    \item We introduce multimodal matching conditioned visual servo policy to overcome the potential errors of explicit matching.
    \item Several architectural designs are proposed, including translation-equivariant probabilitic matching representation, scale and intrinsic decoupled velocity to ease the training difficulties.
    \item We validate CNSv2 both in simulation and real-world, demonstrating its ability to handle textureless scenes and illumination variations, while preserving the generalization and real-time ability of CNS.
\end{itemize}

%===============================================================================
\section{Related Works}

\textbf{Visual Servo.} Traditional visual servo includes IBVS, PBVS and hybrid approaches. IBVS derives the Jacobian of matched keypoint positions versus to camera velocity for control, and is robust to calibration errors but face Jacobian singularities, local minima \cite{chaumette2007potential}. PBVS uses camera's 3D poses as features and is globally asymptotically stable. However, estimating camera's pose also requires explicit correspondence for solving epipolar constraint or PnP problem, which is sensitive to the errors of camera intrinsic and 3D model. It may also lead to unsatisfactory image trajectory that features leave the camera field of view \cite{chaumette2006visual}. Hybrid approaches switch between \cite{gans2007switch} or combine \cite{kermorgant2011combine,malis19992} the two methods to utilize the both advantages.

Recent learning based methods improve the traditional methods in three ways. (1) The first one \cite{adrian2022dfbvs} tries to improve the quality of explicit correspondence using recent learning based image matching methods or optical-flow estimation methods. (2) The second one focuses on improving the controller or the both. \cite{puang2020kovis} trains scene-specific neural observer to predict accurate keypoints for pose-specific neural controller. \cite{chen2024cns} models the matched keypoints as graph and utilize graph neural networks to handle arbitrary number of keypoints and desired pose. (3) The last one \cite{saxena2017exploring,bateux2018training,yu2019siamese,felton2021siame} adopts the implicit feature for control, achieving comparable precision with classic methods and is robust to feature error, however, facing challenges in generalization.

\textbf{Image Matching.} Traditional image matching follows three steps: keypoint detection, keypoint description and descriptor matching. Handcrafted critics \cite{bay2006surf,lowe2004distinctive} or deep neural networks \cite{savinov2017quad,ebel2019beyond,mishchuk2017working,tian2017l2} are introduced for keypoint detection or description.  
These detector-based methods generate sparse matches and tend to fail in textureless scenes as there's no feature points to be extracted. 

Recent detector-free approaches replace the keypoint detection with dense feature matching in coarsest level, followed by several local refine modules to predict dense or semi-dense matches. LoFTR \cite{sun2021loftr} is the first to utilize the Transformer for detector-free matching, effectively capturing long-range dependencies. Efficient LoFTR \cite{wang2024efficient} addresses computational cost of densely transforming entire coarse feature maps by reducing redundant computations through adaptive token selection. DKM \cite{edstedt2023dkm} adopts a dense matching approach, departing from the sparse paradigm by refining matches through stacked feature maps and depthwise convolution kernels. RoMa \cite{edstedt2024RoMa} leverages frozen pretrained features from foundation models to achieve robust matching.

\textbf{Foundation Vision Model.} Foundation vision models pretrained on large quantities of data aim to provide features for universal downstream vision tasks. DINO \cite{caron2021dino} works by interpreting self-supervision as a special case of self-distillation, the resulting features are more distinctive than that trained by supervision. With training on sufficient data, DINOv2 \cite{oquab2023dinov2} generate visual features that are robust and perform well across domains without any requirement for fine-tuning. AM-RADIO \cite{ranzinger2024radio} distills large vision foundation models (including CLIP \cite{radford2021clip} variants, DINOv2 \cite{oquab2023dinov2}, and SAM \cite{kirillov2023sam}) into a single one, serving as a superior replacement for vision backbones.

%===============================================================================

\section{Methods}
\label{sec:methods}

\begin{figure*}[t]
\includegraphics[width=0.9\linewidth]{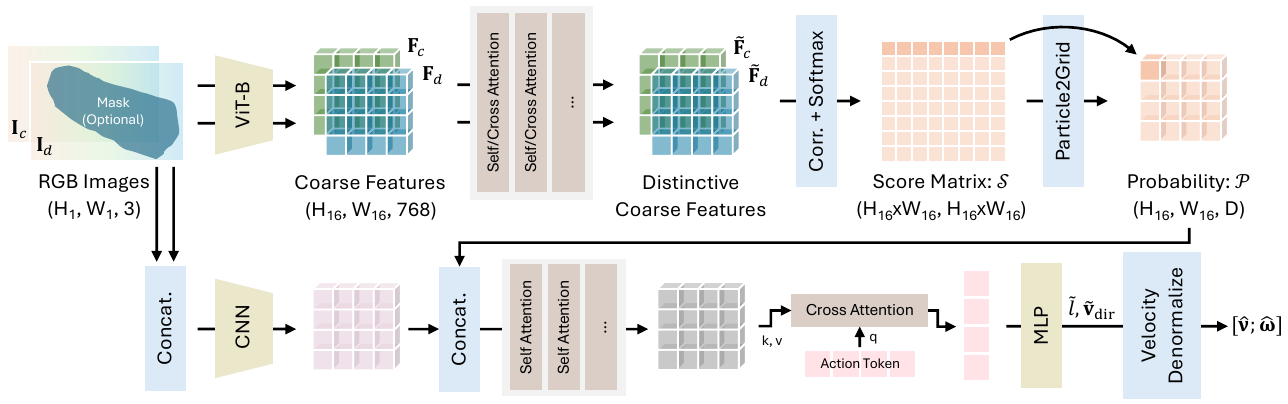}
\centering
\caption{Overview of probabilistic matching conditioned neural policy. We use foundation vision models to extract robust coarse features for matching. We build resolution-agnostic and translation-equivariant representation of probabilistic matching, on which the neural controller is conditioned to predict the velocity control. Fine-grained features from CNN are also fused to capture the pixel-wise error to improve the servo precision.}
\label{fig: overview}
  \vspace{-0.5cm}
\end{figure*}

Given a desired image $\mathbf{I}_d$ and current observed image $\mathbf{I}_c$, our objective is to calculate velocity control that guides the robot to the desired pose, making the two images consistent. Traditional matching based methods fails on textureless scenes and is sensitive to large illumination or view-point changes. We tackle this problem with probabilistic representation of feature matching from foundation models and empower control with data. An overview of method is presented in Fig. \ref{fig: overview}. We will first briefly introduce the traditional pipeline of position-based visual servo and introduce our neural network based policy afterwards. 

\subsection{Preliminaries}

\textbf{Image Matching.} Given a pair of images $\{\mathbf{I}_c, \mathbf{I}_d\}$, detector based matching methods \cite{revaud2019r2d2,detone2018superpoint,sarlin2020superglue} first detects two sets of keypoints $\{ \mathbf{x}_c, \mathbf{x}_d \}$ and extract keypoint descriptors $\{ \mathbf{d}_c, \mathbf{d}_d \}$, then matches are derived from sets of keypoints with minimum mutual descriptor distance. Recent detector free matching methods \cite{sun2021loftr,edstedt2023dkm,edstedt2024RoMa,wang2024efficient} estimate dense or semi-dense matches according to the correlations of image features. We denote match as $\mathcal{M}_i = \{ \mathbf{x}_c^i, \mathbf{x}_{c \rightarrow d}^i \}$. Optical flow estimation methods can also be regarded as dense image matching as they estimate the relative 2D motion vector $\mathcal{F}$ for each pixel in image, and the match can be written as $\mathcal{M}_i = \{ \mathbf{x}_c^i, \mathbf{x}_c^i + \mathcal{F}_i \}$.

\textbf{Epipolar Geometry.} Solving relative transformation from matched keypoints is known as epipolar geometry. 
Given current camera pose $\leftindex^w_c{\mathbf{T}}$ and desired camera pose $\leftindex^w_d{\mathbf{T}}$ with pinhole camera intrinsic $\mathbf{K}$, two matched points in the image plane of each camera as $\mathbf{p}_c = [u_c, v_c, 1]^\top$, $\mathbf{p}_d = [u_d, v_d, 1]^\top$, the positions of two points in normalized camera plane are $\mathbf{x}_c = \mathbf{K}^{-1} \mathbf{p}_c$, $\mathbf{x}_d = \mathbf{K}^{-1} \mathbf{p}_d$, respectively. Denote the 3D position of the point in current camera coordinate as $\leftindex^c{\mathbf{P}} = [X_c, Y_c, Z_c]^\top$ and in desired camera coordinate as $\leftindex^d{\mathbf{P}} = [X_d, Y_d, Z_d]^\top$, we will have:

\begin{equation}
    \label{eq: epipolar1}
    Z_c \mathbf{p}_c = \mathbf{K} \leftindex^{c}{\mathbf{P}}, \quad Z_d \mathbf{p}_d = \mathbf{K} ( \leftindex^d_c {\mathbf{R}} \leftindex^c {\mathbf{P}} + \leftindex^d{\mathbf{t}}_c)
\end{equation}

Eq. \ref{eq: epipolar1} can be transformed into a more compact formula:
\begin{equation}
    \label{eq: epipolar2}
    % \mathbf{x}_{d}^\top \left[ \leftindex^d{\mathbf{t}}_c \right]_{\times} \leftindex^d_c {\mathbf{R}} \mathbf{x}_c = 0
    \mathbf{x}_{d}^\top \leftindex^d{\mathbf{t}}_c^{\wedge} \leftindex^d_c {\mathbf{R}} \mathbf{x}_c = 0
    % \mathbf{x}_{d}^\top \left[ \mathbf{t} \right] ^{d}_{c} \mathbf{R} \mathbf{x}_c = 0
\end{equation}
In Eq. \ref{eq: epipolar2}, $\mathbf{E} = \leftindex^d{\mathbf{t}}_c^{\wedge}  \leftindex^d_c {\mathbf{R}}$ is known as essential matrix, which can be estimated via 5-points or 8-points method. Denote the SVD decomposition of $\mathbf{E}$ as:
\begin{equation}
    \mathbf{E} = \mathbf{U} \mathbf{\Sigma} \mathbf{V}^\top
\end{equation}
The rotation and translation can be recovered from one of the followings that satisfies the positive depth constraint:
\begin{equation}
    \label{eq: epipolar3}
    \begin{aligned}
        \leftindex^d{\mathbf{t}}_{c,1}^{\wedge}  &= \mathbf{U} \mathbf{R}_z \!\left(\frac{\pi}{2}\right) \mathbf{\Sigma} \mathbf{U}^\top, & \leftindex^d_c{\mathbf{R}} _1 &= \mathbf{U} \mathbf{R}_z^\top \!\left( \frac{\pi}{2} \right) \mathbf{V}^\top \\
        \leftindex^d{\mathbf{t}}_{c,2}^{\wedge}  &= \mathbf{U} \mathbf{R}_z \!\left( -\frac{\pi}{2} \right) \mathbf{\Sigma} \mathbf{U}^\top, & \leftindex^d_c{\mathbf{R}} _2 &= \mathbf{U} \mathbf{R}_z^\top \!\left(-\frac{\pi}{2} \right) \mathbf{V}^\top
    \end{aligned}
\end{equation}

\textbf{Position-Based Visual Servo}. Given relative transformation $\leftindex^d_c{\mathbf{R}}, \leftindex^d{\mathbf{t}}_c$ from current pose to desired pose, a camera velocity control scheme can be established to decay the pose error exponentially to zero:
\begin{equation}
\label{eq: PBVS_straight}
    \leftindex^c{\bm \upnu}_c = - \lambda \leftindex^d_c{\mathbf{R}}^\top \leftindex^d{\mathbf{t}}_c, \quad \leftindex^c{\bm \upomega}_c = - \lambda \theta \mathbf{u}
\end{equation}
where $\theta \mathbf{u}$ is the axis-angle representation of $\leftindex^d_c{\mathbf{R}}$, and $\lambda$ controls the error decay speed. If pose involved in Eq. \ref{eq: PBVS_straight} is perfectly estimated, the camera trajectory would be a straight line in Cartesian space. However, the image trajectory may not be satisfactory enough, because in same particular configurations some important visual cues may leave the camera field of view \cite{chaumette2006visual}.

\subsection{From Traditional Pipeline to Neural Policy}

In summary, traditional visual servo from image pairs with position-based control is in three parts: 
\begin{enumerate} [leftmargin=*]
    \item An image matcher giving sets of explicit correspondence: 
    \begin{equation}
        \mathcal{M} \gets {\rm ImageMatcher}(\mathbf{I}_c, \mathbf{I}_d)
    \end{equation}
    \item Relative pose estimation via solving epipolar constraint: 
    \begin{equation}
    \label{eq: epipolar_solver}
        \{ \leftindex^d_c{\mathbf{R}}, \leftindex^d{\mathbf{t}}_c \} \gets {\rm EpipolarSolver}(\mathcal{M})
    \end{equation}
    \item Position-based control law to move camera: 
    \begin{equation}
    \label{eq: pbvs_solver}
        \{ \leftindex^c{\bm \upnu}_c, \leftindex^c{\bm \upomega}_c \} \gets {\rm PBVS}( \leftindex^d_c{\mathbf{R}}, \leftindex^d{\mathbf{t}}_c )
    \end{equation}
\end{enumerate}

Our neural policy borrows the traditional pipeline and is also in three steps:
\begin{enumerate} [leftmargin=*]
    \item A patch matcher giving dense probabilistic matching scores of coarse features $\mathbf{F}$ from foundation models:
    \begin{equation}
        \mathcal{S} \gets {\rm PatchMatcher}(\mathbf{F}_c, \mathbf{F}_d)
    \end{equation}
    \item A transformer based controller regressing the normalized camera velocity control in unit world and canonical camera configurations:
    \begin{equation}
        \leftindex^c{\tilde{\mathbf{v}}}_c \gets {\rm NeuralController}(\mathbf{F}_c, \mathcal{S})
    \end{equation}
    \item An analytical velocity transform scheme to denormalize camera velocity with real-world camera intrinsic $\mathbf{K}_{\rm real}$ and scene scale $d^*$:
    \begin{equation}
        \leftindex^c{\hat{\mathbf{v}}}_c \gets {\rm VelocityDenormalizer}(\leftindex^c{\tilde{\mathbf{v}}}_c, \mathbf{K}_{\rm real}, d^*)
    \end{equation}
\end{enumerate}

\subsection{Features for Matching}
Foundation vision models provide semantic features robust to illumination and viewpoint changes. We use AM-RADIOv2.5 \cite{ranzinger2024radio} with ViT-B \cite{dosovitskiy2020vit} structure of patch size 16 to extract coarse feature maps $\mathbf{F} \in \mathbb{R}^{H_{16}\times W_{16}\times 768}$ of images $\mathbf{I} \in \mathbb{R}^{H_1 \times W_1 \times 3}$ ($H_{16} = H_1 / 16, W_{16} = W_1 / 16$) :
\begin{equation}
    \mathbf{F}_c = {\rm ViT} (\mathbf{I}_c), \quad \mathbf{F}_d = {\rm ViT} (\mathbf{I}_d)
\end{equation}
Following the practices of LoFTR \cite{sun2021loftr} and GMFlow \cite{xu2022gmflow}, we add a transformer with several self/cross attention layers to make features more distinctive for matching:
\begin{equation}
    \{ \tilde{\mathbf{F}}_c, \tilde{\mathbf{F}}_d \} = {\rm Transformer}(\mathbf{F}_c, \mathbf{F}_d)
\end{equation}
We use 2D axial \cite{heo2024ropevit} RoPE \cite{su2024roformer} for positional encoding to better generalization to different image resolution.

\subsection{Probabilistic Matching Representation}
Dense feature matching scores can be computed from correlations of flattened coarse features $\bar{\mathbf{F}} \in \mathbb{R}^{(H_{16}\times W_{16})\times C}$:
\begin{equation}
    \mathcal{S}_{c \rightarrow d} = {\rm softmax}\left( \frac{  \bar{\mathbf{F}}_c \cdot {\bar{\mathbf{F}}_d}^\top  }{\sqrt{C}}  \right)
\end{equation}
As shown in Fig. \ref{fig: overview}, each row of score matrix $\mathcal{S}$ represents the matching distribution between the current patch (with coordinates $\mathbf{x}_c^i$, $i \in \{1,2,..., H_{16}\times W_{16}\}$) and all the patches of desired image. To obtain the explicit correspondence, we can use weighted sum of patch coordinates of desired image: 
\begin{equation}
    \mathbf{x}_{c \rightarrow d}^i = \sum_{j=1}^{ H_{16}\times W_{16} } \mathcal{S}_{c \rightarrow d}^{i,j} \cdot \mathbf{x}_d^j
\end{equation}
For image matching tasks, the coarse match would be $\mathcal{M}_i = \{ \mathbf{x}_c^i, \mathbf{x}_{c \rightarrow d}^i \}$, whereas for flow estimation tasks, the coarse flow would be $\mathcal{F}_i = \mathbf{x}_{c \rightarrow d}^i - \mathbf{x}_c^i$. The explicit correspondence representation reduce the matching distribution into 2 dimensions via WeightedSum operator, thus losing the multimodalities. Our policy is conditioned on the score matrix $\mathcal{S}$ and we don't make any assumption on the reduction operator, which preserves the multimodalities until the last layer that projects the features of matching distribution into 6-dimensional velocity.

However, $\mathcal{S}$ cannot be regarded as features directly, as its channel size (the last dimension) depends on the resolution of input images. Moreover, it is not translation-equivariant, meaning if both the matched patches shift in the spatial dimension, the resulting $\mathcal{S}$ not only shifts in spatial dimension but also shuffles in the channel dimension, which increases the learning difficulty. To ease this, we build a resolution-agnostic and translation-equivariant probabilistic matching representation from score matrix. Inspired from the particle-to-grid operation used in material-point-method \cite{hu2018mpm} that projects the particle's quantity to predefined grids, we project the matching score to predefined anchors. Here, the particles are entries of $\mathcal{S}$ with position $\mathbf{f}_{c}^{i,j} = \mathbf{x}_d^j - \mathbf{x}_c^i$ and quantity $\mathcal{S}_{c \rightarrow d}^{i,j}$. We generate total $D = K \times K$ grid anchors with grid size $\mathbf{g} = [g_x, g_y] = [\frac{2W_{16}}{K}, \frac{2H_{16}}{K}]$. The position of anchors $\mathbf{x}_a$ ranges from $[-W_{16}, -H_{16}]$ to $[W_{16}, H_{16}]$, covering all the possible value ranges of $\mathbf{f}_c$. For each anchor, it's value is accumulated from nearby particles:
\begin{equation}
    \mathcal{P}_{h,w,j} = \frac{ \sum_{k \in \mathcal{N}( \mathbf{x}_a^j )} \mathcal{S}_{c \rightarrow d}^{i,k} \cdot \mathcal{K} \left( \frac{\mathbf{f}_c^{i,k} - \mathbf{x}_a^j}{\mathbf{g}} \right) }{ \sum_{k \in \mathcal{N}( \mathbf{x}_a^j )} \mathcal{K}\left( \frac{\mathbf{f}_c^{i,k} - \mathbf{x}_a^j}{\mathbf{g}} \right) }
\end{equation}
where $i = h \times W_{16} + w \in \{1, 2, ..., H_{16} \times W_{16}\}$, $j \in \{1, 2, ..., K \times K\}$. $\mathcal{N}( \mathbf{x}_a^i )$ finds all the particles within the searching radius (=$1.5 \mathbf{g}$) of anchor grid $i$, and $\mathcal{K}$ is 2D quadratic B-spline kernel:
\begin{equation}
    \begin{aligned}
        & \mathcal{K}(\mathbf{a}) = \kappa(a_x) \cdot \kappa(a_y) \\
        & \kappa(a) = \left\{ 
        \begin{aligned}
            % &\frac{3}{4} - \lVert a \rVert^2, & 0 \leq \lVert a \rVert \le \frac{1}{2} \\
            % &\frac{1}{2} \left( \frac{3}{2} - \lVert a \rVert^2 \right)^2, & \frac{1}{2} \leq \lVert a \rVert \le \frac{3}{2}
            &0.75 - \lvert a \rvert^2, & 0 & \leqslant \lvert a \rvert < 0.5 \\
            &0.5 \left( 1.5 - \lvert a \rvert \right)^2, & 0.5 & \leqslant \lvert a \rvert < 1.5 \\
            &0, & 1.5 & \leqslant \lvert a \rvert
        \end{aligned} 
        \right.
    \end{aligned}
\end{equation}
Ablation study (Fig. \ref{fig: trans_equiv}) shows this resolution-agnostic and translation-equivariant representation of matching probability helps the network converge faster than using score matrix.

\subsection{Velocity Denormalization}
It is unrealistic to cover all the possible distributions of camera intrinsics and scene scales in training data. Instead, we train neural policy with data generated from unit world with a canonical camera instrinsic to predict normalized velocity $\tilde{\mathbf{v}} = [\leftindex^c{\tilde{\bm \upnu}}_c; \leftindex^c{\tilde{\bm \upomega}}_c ]$, and try to find a mapping that denormalizes the velocity to real-world configurations.

We first recover the normalized relative pose from the normalized velocity (reverse the process of Eq. \ref{eq: PBVS_straight}):
\begin{equation}
\label{eq: inv_PBVS}
    \{ \leftindex^d_c{\tilde{\mathbf{R}}},  \leftindex^d{\tilde{\mathbf{t}}}_c   \} \gets {\rm PBVS}^{-1}( \leftindex^c{\tilde{\bm \upnu}}_c,  \leftindex^c{\tilde{\bm \upomega}}_c )
\end{equation}

\textbf{Adapt to Real-World Scene Scale.} Suppose we have a well trained neural controller that estimates perfect normalized velocity from probabilistic matching and according to Eq. \ref{eq: epipolar1}, we would have:
\begin{equation}
    \label{eq: rewrite ep1}
    \tilde{Z}_d \mathbf{K}^{-1} \mathbf{p}_d = \tilde{Z}_c \leftindex^d_c{\tilde{\mathbf{R}}} \mathbf{K}^{-1} \mathbf{p}_c + \leftindex^d{\tilde{\mathbf{t}}}_c
\end{equation}
If the real depth is $s$ times of that in training (in the unit world): $\hat{Z}_d = s \tilde{Z}_d$, $\hat{Z}_c = s \tilde{Z}_c$, then the solution of Eq. \ref{eq: rewrite ep1} would be $\leftindex^d_c{\hat{\mathbf{R}}} = \leftindex^d_c{\tilde{\mathbf{R}}}$, $\leftindex^d{\hat{\mathbf{t}}}_c = s \leftindex^d{\tilde{\mathbf{t}}}_c$. Control using pose errors in real-world scale with PBVS yields:
\begin{equation}
    \leftindex^c{\hat{\bm \upnu}}_c = s \leftindex^c{\tilde{\bm \upnu}}_c, \quad \leftindex^c{\hat{\bm \upomega}}_c = \leftindex^c{\tilde{\bm \upomega}}_c
\end{equation}
In the unit world, we assume $\tilde{Z}_d = 1$ and in real-world, we have $s = d^*$.

\textbf{Adapt to Real-World Pinhole Camera.} 
If the actual focal length is $s$ times of that used in training: $\hat{f} = s\cdot \tilde{f}$ and suppose $(c_x, c_y)$ are always exactly the half of image width and height, we would have $\hat{\mathbf{x}} = \mathbf{S}^{-1} \tilde{\mathbf{x}} $ where $\mathbf{S} = {\rm diag}[s, s, 1]$. According to Eq. \ref{eq: epipolar2}, we would have:
\begin{equation}
    \label{eq: epipolar4}
    \hat{\mathbf{x}}_{d}^\top \mathbf{S}^\top \leftindex^d{\tilde{{\mathbf{t}}}}_c^{\wedge} \leftindex^d_c {\tilde{\mathbf{R}}} \mathbf{S} \hat{\mathbf{x}}_c = 0
\end{equation}
Solving Eq. \ref{eq: epipolar4} gives $\hat{\mathbf{E}} = \mathbf{S}^\top \leftindex^d{\tilde{\mathbf{t}}}_c^{\wedge} \leftindex^d_c {\tilde{\mathbf{R}}} \mathbf{S}$ where $\leftindex^d{\tilde{\mathbf{t}}}_c$ and $\leftindex^d_c{\tilde{\mathbf{R}}}$ are already known according to Eq. \ref{eq: inv_PBVS}. Therefore, we could decompose $\hat{\mathbf{E}}$ with SVD again and follow the steps in Eq.\ref{eq: epipolar3} and Eq. \ref{eq: PBVS_straight} to get the right velocity control under real-world configurations. %and select the solution satisfying the positive depth constraint to give the right prediction under actual camera intrinsic configurations.

\subsection{Supervision}
The neural controller predicts the log-norm $\tilde{l}$ and direction $\tilde{\mathbf{v}}_{\rm dir}$ of normalized velocity:
\begin{equation}
    \tilde{\mathbf{v}} = \sigma \! \left(\tilde{l}\right) \cdot \frac{\tilde{\mathbf{v}}_{\rm dir}}{\norm{\tilde{\mathbf{v}}_{\rm dir}}}, \quad \sigma(x) = \left\{
    \begin{aligned}
        & e^{x - 1}, & x \leqslant 1 \\
        & x, & x > 1
    \end{aligned}
    \right.
\end{equation}
We use L1-loss to supervise the log norm and use cosine similarity loss to supervise the direction:
\begin{equation}
    \mathcal{L}_{\rm norm} = \lvert \ \sigma^{-1}\left(  \norm{ \tilde{\mathbf{v}}^* } \right) - \tilde{l} \ \rvert, \quad
    \mathcal{L}_{\rm dir} = 1 - \frac{ \tilde{\mathbf{v}}^* \cdot \tilde{\mathbf{v}}  }{ \norm{\tilde{\mathbf{v}}^*} \norm{ \tilde{\mathbf{v}} } }
\end{equation}
where $\tilde{\mathbf{v}}^*$ is the ground truth normalized camera velocity computed from ground truth relative pose using PBVS.

\subsection{Safe Velocity Control}
\begin{figure}[t]
\includegraphics[width=\linewidth]{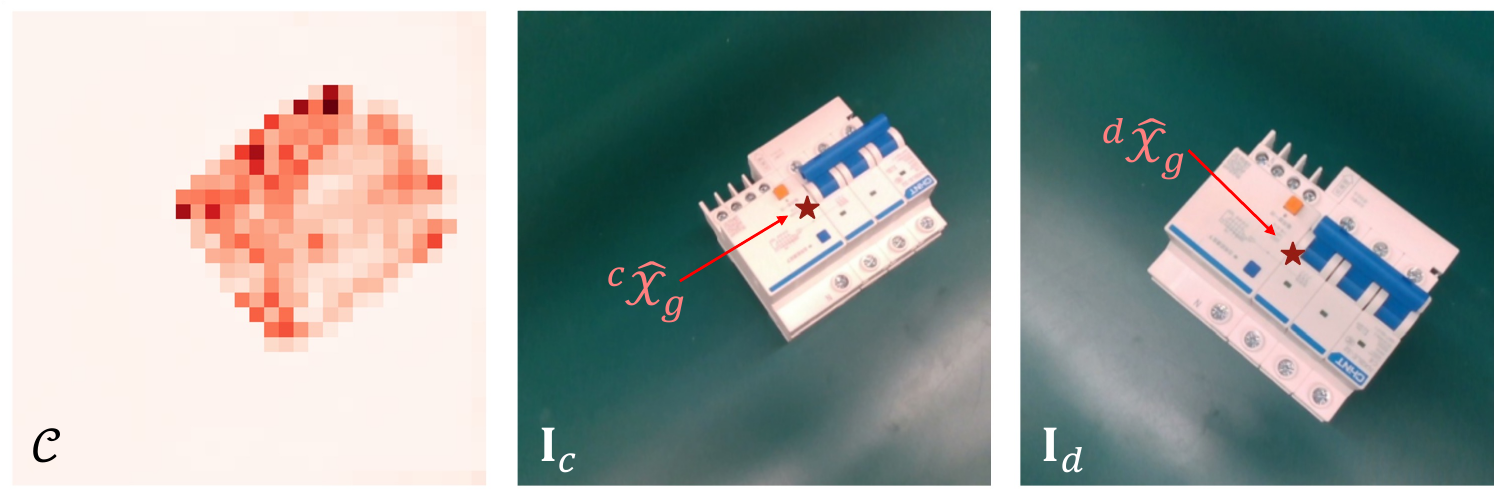}
\centering
% \vspace{-0.5cm}
\caption{An example of dual softmax matching score $\mathcal{C}$ (darker means higher confidence) and estimated image gravity centers of image pairs in real-world experiments. The estimated gravity centers may not be very accurate but are good enough for hybrid velocity control in the early stage of servo. }
\label{fig: conf_gravity}
  \vspace{-0.5cm}
\end{figure}

PBVS with Eq. \ref{eq: PBVS_straight} may face feature loss problem \cite{chaumette2006visual}. With additional coarse correspondence, a hybrid visual servo scheme can be established to realize both object-centric image trajectory and straight Cartesian trajectory. % keep the servo object in camera field of view and still move straight in Cartesian space: 
\begin{equation}
    \mathbf{v} = -\lambda \hat{\mathbf{J}}^{-1} \mathbf{e}, \quad \mathbf{e} = \left[ \leftindex^d{\hat{\mathbf{t}}}_c; \leftindex^c{\hat{\mathcal{X}}}_g - \leftindex^d{\hat{\mathcal{X}}}_g; \hat{\theta} \hat{\mathbf{u}}_z \right]
\end{equation}
where $\mathbf{J}$ is the Jacobian of error versus velocity, we would suggest referring to \cite{malis19992} for details. 
% $\hat{\theta} \hat{\mathbf{u}}_z$ is the last item of axis-angle representation of $\leftindex^d_c{\hat{\mathbf{R}}}$. 
$\mathcal{X}_g$ is the gravity center of an image, we estimate it as the weighted (using dual softmax matching scores $\mathcal{C}^i = \sum_{j=1}^{N_{16}} \mathcal{S}_{c \rightarrow d}^{i, j} \cdot \mathcal{S}_{d \rightarrow c}^{i, j}, N_{16} = H_{16} \times W_{16}$) sum of patch coordinates (example in Fig. \ref{fig: conf_gravity}):
\begin{equation}
    % \begin{aligned}
        \leftindex^c{\hat{\mathcal{X}}}_g = \frac{ \sum_{i=1}^{N_{16}} \mathcal{C}^i \cdot \mathbf{x}_c^i }{ \sum_{i=1}^{N_{16}} \mathcal{C}^i }, \quad 
        \leftindex^d{\hat{\mathcal{X}}}_g = \frac{ \sum_{i=1}^{N_{16}} \mathcal{C}^i \cdot ( \mathbf{x}_c^i + \mathcal{F}_i ) }{ \sum_{i=1}^{N_{16}} \mathcal{C}^i } 
    % \end{aligned}
\end{equation}
As $\mathbf{J}$ can be computed from $\leftindex^d_c{\hat{\mathbf{R}}}, \leftindex^d{\hat{\mathbf{t}}}_c, \leftindex^c{\hat{\mathcal{X}}}_g$, we choose to use this hybrid control when $\norm{ \leftindex^c{\hat{\mathcal{X}}}_g - \leftindex^d{\hat{\mathcal{X}}}_g } > 0.1 \sqrt{N_{16}}$ to achieve both satisfactory Cartesian and image trajectories, and switch to PBVS control (which is directly supervised) for higher precision when close to the desired pose.

\subsection{Training Details}

\begin{figure}[tb]
\includegraphics[width=0.9\linewidth]{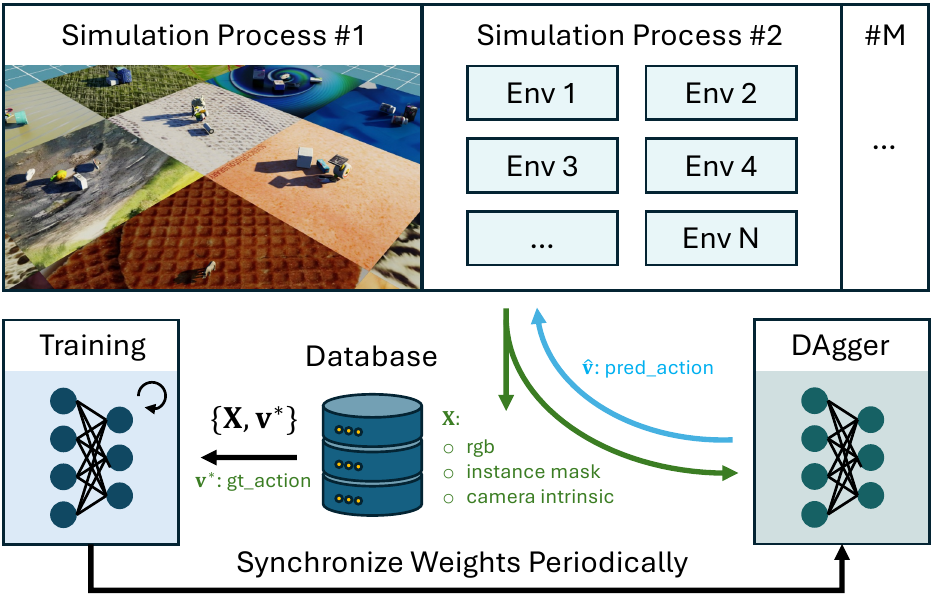}
\centering
\caption{Training pipeline. We use NVIDIA IsaacSim to render photo-realistic images. One simulation process collects data from uniformly sampled space. Another simulation process collects data with DAgger.}
\label{fig: dagger}
  % \vspace{-0.2cm}
\end{figure}

\begin{figure}[tb]
\includegraphics[width=\linewidth]{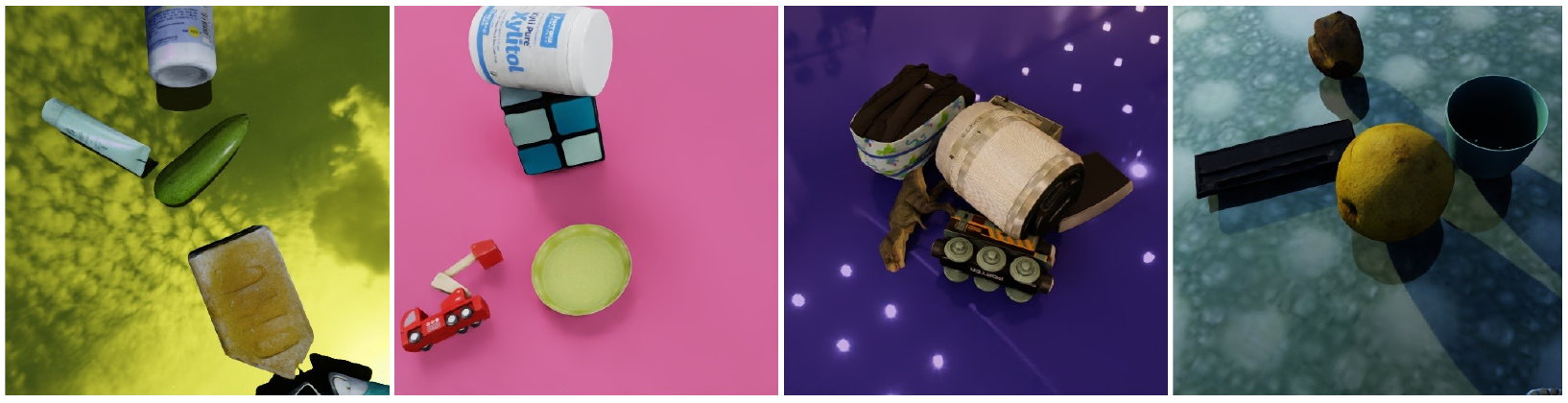}
\centering
\caption{Examples of rendered images in simulation environment.}
\label{fig: render}
  % \vspace{-0.2cm}
\end{figure}

Our training data is generated purely in simulation. We use IsaacSim to render realistic images. We launch two sampling processes, one uniformly samples current and desired pose pairs for rendering in the upper hemisphere offline; Another uniformly samples the initial and desired poses, and adopts the DAgger \cite{ross2011dagger} scheme which updates current poses online for rendering with actions from current training neural policy.

We use 3D models from GSO \cite{downs2022gso} and OmniObject3D \cite{wu2023omniobject3d} datasets (total 6852 models). We randomly scatter 1$\sim$6 objects on the ground plane as a servo scene. Randomization domains include object sizes and poses, background textures (32k images) and materials, and ambient lights (733 HDR maps). Example rendered images are shown in Fig. \ref{fig: render}.

\section{Experimental Results}
\label{sec:result}

\subsection{Translation-Equivariance Makes Training Easy}

\begin{figure}[t]
\includegraphics[width=\linewidth]{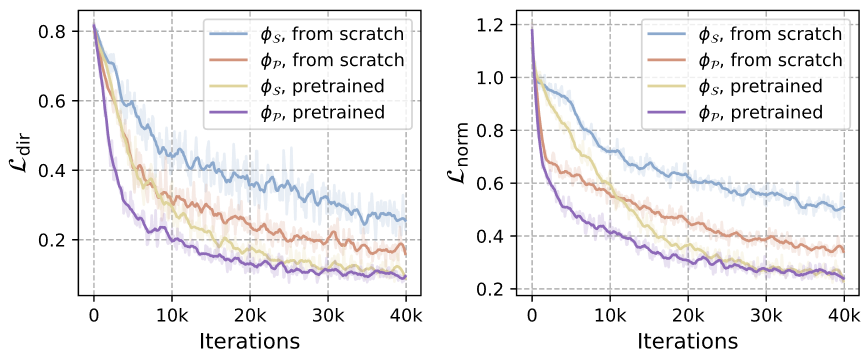}
% \includesvg[width=\linewidth]{figs/ablation_on_P.svg}
\centering
\vspace{-0.6cm}
% \caption{Improvement of training convergence when adopting the translation-equivariant probabilistic matching representation.}
\caption{Translation-equivariant probabilistic matching representation enables faster convergence of training.}
\label{fig: trans_equiv}
  \vspace{-0.2cm}
\end{figure}

% We train
We train our controller conditioned on raw score matrix (denote as model $\phi_{\mathcal{S}}$) and our translation-equivariant representation (model $\phi_{\mathcal{P}}$) respectively. Note to make $\phi_{\mathcal{S}}$ also resolution-agnostic, we bilinearly sample the last dimension of $\mathcal{S}$ from size $H_{16}\times W_{16}$ to $K \times K$. Fig. \ref{fig: trans_equiv} shows the first 40k iterations (with batch size of 16) training loss, $\phi_{\mathcal{P}}$ converges faster than $\phi_{\mathcal{S}}$, which is more obvious when training the image backbone from scratch rather than using foundation models. This shows the superiority of our translation-equivariant probabilistic matching representation.

\subsection{Effectiveness of Architecture Designs}

% \begin{table}[htbp]
\begin{table}[tb]
\centering
\caption{Performance of different network architectures and environment configurations in simulation.}
\label{tab: ablation_arch}
\resizebox{\linewidth}{!}{

\begin{tabular}{l|l|l|l}
\hline
Configurations                                          & SR         & TE (mm)       & RE ({\textdegree})  \\ \hline
(1) Baseline                                            & 20/20      & 0.948±0.606 & 0.075±0.048           \\ \hline
(2) $\phi_{\mathcal{P}} \rightarrow \phi_{\mathcal{M}}$ & 19/20      & 3.517±2.977 & 0.301±0.254          \\ \hline
(3) $f = 256, d^* = 0.5{\rm m}$, aware                  & 19/20      & 0.324±0.391 & 0.057±0.060          \\ \hline
(4) $f = 768, d^* = 1.5{\rm m}$, aware                  & 20/20      & 1.508±1.202 & 0.080±0.065          \\ \hline
(5) $f = 256, d^* = 0.5{\rm m}$, unaware                & 16/20      & 0.291±0.323 & 0.046±0.045          \\ \hline
(6) $f = 768, d^* = 1.5{\rm m}$, unaware                & 20/20      & 1.140±0.738 & 0.066±0.041          \\ \hline
(7) Hybrid $\rightarrow$ PBVS                           & 18/20      & 1.064±0.655 & 0.085±0.056           \\ \hline
\end{tabular}
}
\vspace{-0.2cm}
\end{table}

We first define metrics for evaluating policies: (1) SR: success ratio of each run; (2) TE: final translation error of each servo episode; (3) RE: final rotation error of each servo episode.

We evaluate our neural policy in simulation with different architectures and different environment configurations. Our baseline model uses $\phi_{\mathcal{P}}$ as neural controller, and adopts the hybrid control at the initial stage of visual servoing and switches to PBVS when two image gravity centers are close to each other. The baseline environment uses a canonical camera with intrinsic $f_x = f_y = 512, \ c_x = c_y = 256, \ H_1 = W_1 = 512$ and scene scale of $d^* = 1{\rm m}$. The performance is listed at the row (1) in Table \ref{tab: ablation_arch}.

\textbf{Probabilistic Beats Explicit.} When we switch the neural controller to $\phi_{\mathcal{M}}$, \emph{i.e.}, use explicit matching for condition, we would see a significant precision drop in TE and RE (row (2) in Table \ref{tab: ablation_arch}), indicating coarse explicit matching is not sufficient for high precision control.

\textbf{Effectiveness of Velocity Denormalization.} We scale the focal length of canonical camera by $0.5\times$ and $1.5\times$, respectively. As this changes camera's FoV, to ensure the observed images are roughly the same across different configurations, we also zoom the scene by the same scale. As shown in row (3)(4) of Table \ref{tab: ablation_arch}, if the policy is aware of changed parameter to denormalize its prediction, the rotation errors in scaled environments are roughly the same as those in baseline environment, whereas the translation errors are linearly increased with scene scale $d^*$. If the policy is unaware of the changes in parameter, it tends to fail with smaller scene scales (row (5)) as the camera moves too aggressive in such scale. Interestingly, in enlarged scene scale, the policy achieves comparable convergence and precision with those in the unit world (row (6)), indicating convergence is insensitive to the underestimated linear velocity, however, it takes longer time to move to the final pose.

\textbf{Hybrid Control Is More Robust.} Row (7) in Table \ref{tab: ablation_arch} shows success rate drops if we always use PBVS control, but the final precision for successful cases are not affected. The failed cases are with large initial viewpoint deviation inducing feature loss problem.

\subsection{Real-World Experiment Setup}
\begin{figure}[t]
\includegraphics[width=\linewidth]{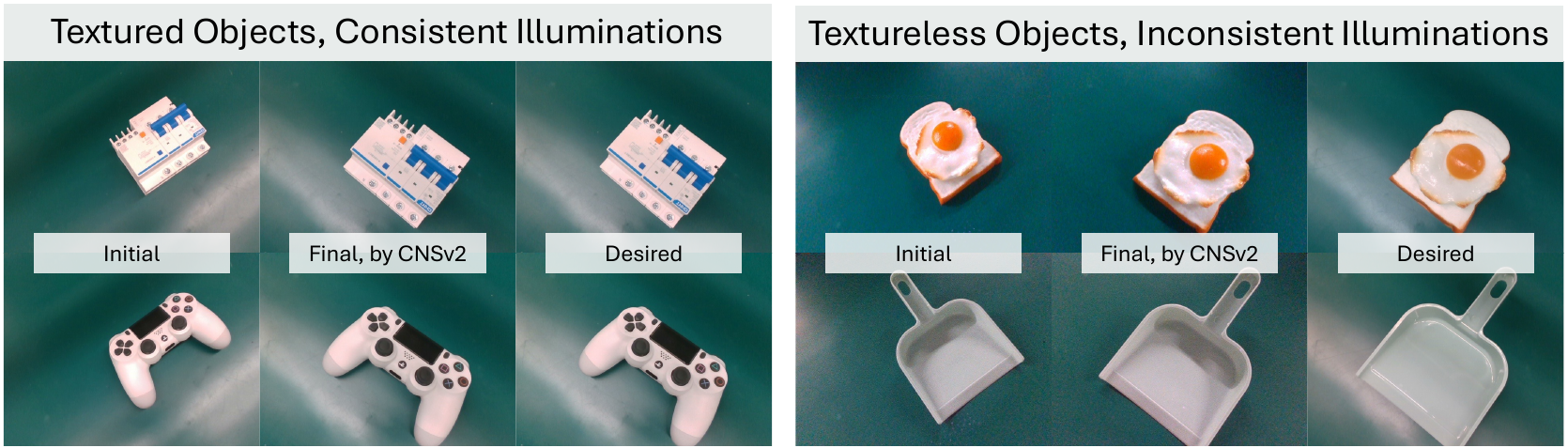}
\centering
% \vspace{-0.5cm}
\caption{Objects and illuminations used in real-world experiments. }
\label{fig: real_exp_scenes}
  % \vspace{-0.4cm}
\end{figure}

We evaluate policies on 4 unseen objects with increasing difficulties (circuit breaker, gamepad, foods and plastic dust pan) and 2 different illumination conditions (Fig. \ref{fig: real_exp_scenes}). The circuit breaker has rich texture and rough surface, whereas the plastic dust pan is textureless and has reflective surface. We use consistent illuminations when sampling desired and initial images for the first two objects but inconsistent illuminations for the last two objects. We sample 10 desired-initial pose pairs for each object. The sampled initial pose rotation errors range from $30.146^\circ$ to $172.127^\circ$ with (mean, std) as $(\mu_{\mathbf{R}} =87.446^\circ, \  \sigma_{\mathbf{R}}=46.593^\circ )$. The sampled initial translation errors range from $62.527 {\rm mm}$ to $265.812 {\rm mm}$ with (mean, std) as $(\mu_{\mathbf{t}}=147.765{\rm mm}, \ \sigma_{\mathbf{t}}=52.193{\rm mm} )$.

Here we define additional metrics for real-world experiments: (1) TT: total convergence time of one servo episode. We use SSIM \cite{wang2004image} to determine the time to stop current episode in easy scenes with consistent illuminations, whereas in hard scenes with inconsistent illuminations, we run all the policies for 30 seconds; (2) FPS: frames per second, which evaluates the neural network's inference speed. Note the time cost of reading camera buffer data is not included. We benchmark all the policies on a RTX 4090 GPU.

\subsection{Real-World Experiment Results}

\begin{table}[t]
  \centering
  \caption{Statistics of real-world experiments. Our model achieves highest success ratio and precision among all the policies and all the scene setups. SIFT-IBVS fails on textureless objects and is sensitive to illumination changes. RoMa+IBVS shows robustness to textureless objects and illumination changes, but tendes to fail when facing large initial in-plane rotations.}
  \label{tab: real_exp_stat}
  \includegraphics[width=\linewidth]{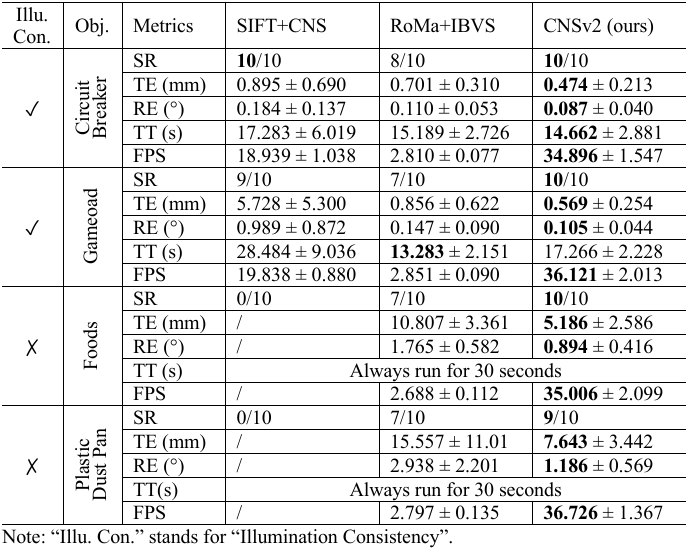}
  \vspace{-0.5cm}
\end{table}

We compare CNSv2 with two methods (Table \ref{tab: real_exp_stat}): (1) SIFT+CNS \cite{chen2024cns}: A graph neural network based controller relying on explicit correspondence from detector-based image matching method, here we use SIFT; (2) RoMa \cite{edstedt2024RoMa} + IBVS: Image servo using explicit matches from the state-of-the-art detector-free dense matching method, RoMa.

Our method achieves highest success ratio and precision in all the scene setups, showing robustness to large in-plane rotation errors and illumination inconsistency and the capability to servo textureless objects. SIFT+CNS is also robust to large in-plane rotation errors (thanks to the rotation invariant nature of SIFT descriptor), but fails on textureless objects and illumination changes. RoMa+IBVS is robust to illumination inconsistency and can handle textureless objects, but the precision is lower. We also find that RoMa+IBVS fails on specific tests having large initial rotation errors.

% \yw{Cases to show heatmap, or probabilistic matching?}

\section{Conclusion}
In this work, we propose CNSv2 that leverages multimodal correspondence as conditions to predict velcoity control for visual servo. We derive the resolution-agnostic and translation-equivariant probabilistic matching representation, together with velocity denomalization technique to ease the training difficulties. 
With elaborate architecture designs, our policy is robust to textureless scenes and illuminiation variations, and generalizes well on novel scenes.

\balance
\bibliographystyle{IEEEtran}
\bibliography{IEEEabrv,icra2025}

\clearpage

\end{document}